\pdfoutput=1

\documentclass[11pt]{article}
\usepackage[final]{ACL2023}

\usepackage{times}
\usepackage{latexsym}
\usepackage{url}
\usepackage{minted}

\usepackage[T1]{fontenc}
\usepackage[utf8]{inputenc}

\usepackage{microtype}
\usepackage{subcaption}
\usepackage{inconsolata}
\usepackage{pifont}
\usepackage{multirow}
\usepackage{booktabs}
\usepackage{todonotes}
\usepackage{xspace}
\usepackage{enumitem}
\usepackage{svg}
\usepackage{amsmath,amssymb}
\usepackage{cleveref}

\newcommand{\note}[4][]{{\todo[author=#2,color=#3,size=\footnotesize,fancyline,caption={},#1]{#4}}}
\definecolor{tticblue}{RGB}{0, 94, 184}

\newcommand{\hala}[2][]{{\note[#1]{hala}{tticblue!20}{#2}}}

\newcommand{\toolname}{\textsc{VLM-Lens}\xspace}
\newcommand{\interalia}[1]{\citep[\textit{inter alia}]{#1}}

\usepackage{graphicx,calc}
\newlength\myheight
\newlength\mydepth
\settototalheight\myheight{Xygp}
\newcommand*\titlegraphics[1]{%
    \settototalheight\myheight{Xygp}%
    \settodepth\mydepth{Xygp}%
    \raisebox{-2.\mydepth}{\includegraphics[height=2\myheight]{#1}}%
}
\newcommand{\logo}{\titlegraphics{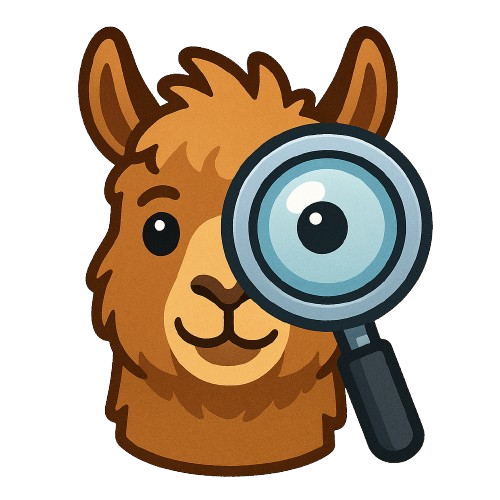}}

\crefname{figure}{Figure}{Figures}
\crefname{table}{Table}{Tables}
\crefname{section}{\S}{\S\S}
\Crefname{section}{\S}{\S\S}

\title{
  From Behavioral Performance to Internal Competence:\\
  Interpreting Vision-Language Models with \textsc{VLM-Lens}\logo
}

\newcommand{\uw}{$^{1}$}
\newcommand{\veci}{$^{2}$}
\newcommand{\um}{$^{3}$}
\newcommand{\sbu}{$^{4}$}
\newcommand{\mcgill}{$^{5}$}
\newcommand{\dc}{$^{6}$}
\newcommand{\andaffto}{$^,$}

\author{
  Hala Sheta\uw\andaffto\veci\andaffto\thanks{\hspace{5pt}Equal contribution.}
  \quad Eric Huang\uw\andaffto\mcgill\andaffto\footnotemark[1]
  \quad Shuyu Wu\um\andaffto\footnotemark[1]
  \\
  \bf Ilia Alenabi\uw
  \; Jiajun Hong\sbu
  \; Ryker Lin\uw
  \; Ruoxi Ning\uw\andaffto\veci
  \; Daniel Wei\uw
  \; Jialin Yang\uw\andaffto\dc \\
  \bf Jiawei Zhou\sbu
  \quad Ziqiao Ma\um
  \quad Freda Shi\uw\andaffto\veci \\
  \uw University of Waterloo \quad \veci Vector Institute \quad \um University of Michigan \\
  \sbu Stony Brook University \quad \mcgill McGill University \quad \dc Dartmouth College \\[2pt]
  \includesvg[width=10pt]{figs/github-mark.svg} \href{https://github.com/compling-wat/vlm-lens}{\texttt{https://github.com/compling-wat/vlm-lens}}
}

\begin{document}
\maketitle
\begin{abstract}
    We introduce \toolname, a toolkit designed to enable systematic benchmarking, analysis, and interpretation of vision-language models (VLMs) by supporting the extraction of intermediate outputs from any layer during the forward pass of open-source VLMs.
    \toolname provides a unified, YAML-configurable interface that abstracts away model-specific complexities and supports user-friendly operation across diverse VLMs.
    It currently supports 16 state-of-the-art base VLMs and their over 30 variants, and is extensible to accommodate new models without changing the core logic.

    The toolkit integrates easily with various interpretability and analysis methods.
    We demonstrate its usage with two simple analytical experiments, revealing systematic differences in the hidden representations of VLMs across layers and target concepts. 
    \toolname is released as an open-sourced project to accelerate community efforts in understanding and improving VLMs.
\end{abstract}

\section{Introduction}

\begin{figure*}[ht!]
    \centering
    \includegraphics[width=1.0\linewidth,page=1]{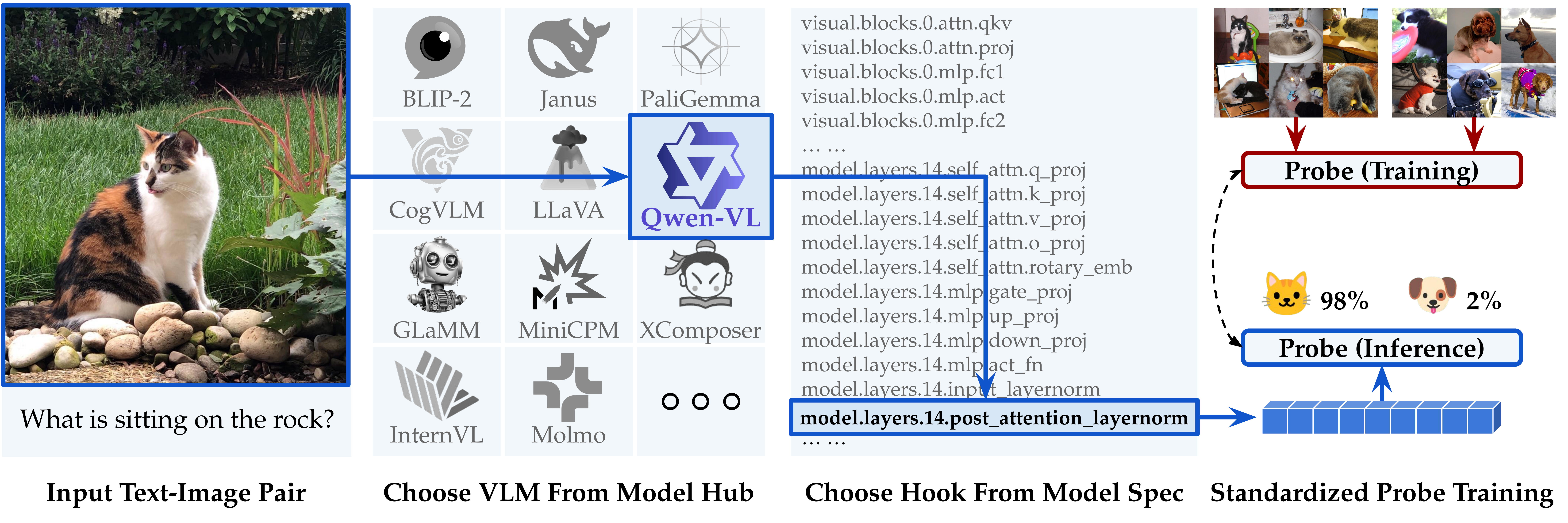}
    \vspace{-20pt}
    \caption{
        An example use case of \toolname, where intermediate output from Qwen2-VL \citep{wang2024qwen2} is extracted for probing.
    }
    \vspace{-10pt}
    \label{fig:teaser}
\end{figure*}

Vision-language models \citep[VLMs;][\textit{inter alia}]{kirillov2023segment, radford2021learning,li2022blipbootstrappinglanguageimagepretraining, liu2023visual, wang2024qwen2} have become essential across a wide range of applications, including multimodal understanding \citep{yue2024mmmumassivemultidisciplinemultimodal}, robotics \citep{li2024embodied}, and world modeling \citep{gao2025vision}.
However, existing VLM benchmarks predominantly adopt exact-match based accuracy and its derivations to evaluate model performance \interalia{lin2014microsoft, johnson2017clevr, yue2024mmmumassivemultidisciplinemultimodal, fu2024mme}, which may either overlook the information embedded in their hidden representations \citep{zhang2025visionlanguagemodelsrepresentspace} or yield misleading assessments due to shortcut exploitation \citep{xu2025overcoming}.
Currently, there lacks a unified framework for extracting the internal representations of VLMs, making it challenging to assess model capabilities that go beyond simple performance evaluations.

Meanwhile, interpretability research and toolkits for VLMs remain underdeveloped compared to their text-only counterparts \interalia{nanda2022transformerlens, belrose2023elicitinglatentpredictionstransformers, ali2025entropylensinformationsignaturetransformer}, posing significant challenges to systematically understanding their internal knowledge and decision-making processes.
To the best of our knowledge, extending existing interpretability toolkits, such as TransformerLens \citep{nanda2022transformerlens}, to support VLMs requires substantial engineering effort, as these tools are primarily designed for text-based Transformers.

To address these challenges in both benchmarking and interpretability, we present \textbf{\toolname} (\cref{fig:teaser}), a toolkit that enables easy extraction of VLM intermediate output from any layer in a forward pass.
The key features include:
\begin{itemize}[leftmargin=*, itemsep=-2pt, topsep=2pt]
    \item \textbf{Unified interface.}
          It abstracts out the model-specific setup and preprocessing complexities, allowing operations across models through a unified interface.
          Users can specify custom configurations via a YAML file with minimal boilerplate code provided, and the toolkit automatically handles model loading, preprocessing, and inference.
    \item \textbf{Model-specific environmental support.}
          Different VLMs often require different, and sometimes mutually conflicting, libraries. 
          To address this issue, \toolname provides model-specific environment setups, each of which can be easily installed with a single-line pip install command.
          A rigorous code review process ensures the consistency and reproducibility of the environment setups across different platforms.
    \item \textbf{Extensive model coverage and flexible nature.}
          The toolkit supports a diverse set of state-of-the-art VLMs, spanning widely used open-source models to recently developed, less-documented ones.
          Currently, \toolname supports 16 base VLMs and over 30 variants, with a highly extensible design that allows users or contributors to add new models with minimal effort.
\end{itemize}

\toolname streamlines analytical tasks for VLMs, such as probing \citep{ettinger-etal-2016-probing}, neural circuit inspection \citep{pmlr-v202-chughtai23a}, and knowledge tracing \citep{basu2024understanding}, as well as diagnosing model capabilities and limitations \citep{zhang2025visionlanguagemodelsrepresentspace, stevens2025sparse}.
As such, we anticipate that the toolkit will enable researchers and practitioners to conduct more fine-grained and rigorously controlled analyses of VLMs.
The toolkit is released under the Apache-2.0 license.

\vspace{-5pt}
\section{Related Work}
\vspace{-5pt}

\vspace{2pt}\noindent\textbf{Vision-language models.}
Since vision and text naturally convey information about the world in two complementary modalities, there has been sustained interest in integrating them within unified frameworks \interalia{kiros2014unifying,radford2021learning,liu2023visual}.
Earlier work primarily encoded images and text in a shared embedding space to facilitate efficient retrieval and matching \citep{kiros2014unifying,faghri2018vse++,radford2021learning}. With recent advances in generative text models, exemplified by \citet{brown2020language}, the focus has shifted toward building large-scale VLMs that generate text conditioned on both images and textual prompts \interalia{liu2023visual,wang2024qwen2,zhang2023internlm}.
While some models offer straightforward hidden-state extraction through open-sourced libraries such as HuggingFace Transformers \citep{wolf2020transformers},\footnote{For example, by setting \texttt{output\_hidden\_state=True} in the forward function of LLaVA-1.5-7B \citep{liu2023visual}; see the documentation at \url{https://huggingface.co/docs/transformers/model_doc/llava}.} many require insufficiently documented customizations.
Additionally, there is no systematic support for extracting intermediate representations beyond the layer-wise output, such as attention maps and intermediate features before layer normalization \citep{ba2016layer}.
\toolname addresses this gap by providing a structured and unified interface for extracting intermediate output across many VLMs, thereby enabling detailed analyses of models.

\vspace{2pt}\noindent\textbf{Performance and competence analysis of VLMs.}
Efforts have been made to benchmark the performance of VLMs on various tasks, such as visual question answering \citep{johnson2017clevr}, image captioning \citep{lin2014microsoft} and general cross-modal understanding \citep{yue2024mmmumassivemultidisciplinemultimodal,fu2024mme}.
These benchmarks largely assess models performance through exact-match based accuracy.
However, accuracies fall short of capturing the full spectrum of model \textit{competence}, which encompasses the internal mechanisms and generalizable knowledge that a model possesses, possibly beyond its observable textual output.
To address this issue, recent work has started to explore the competence of VLMs through more fine-grained analyses via hidden states \citep{stevens2025sparse} or output probability \citep{zhang2025visionlanguagemodelsrepresentspace}.
As such, \toolname offers a toolkit for localizing hidden causal mechanisms in VLM, enabling convenient and flexible model competence analysis and assessing beyond simple accuracy-based evaluations.

\iffalse
    These benchmarks assess model performance and by proxy, its \textit{competence}, largely through analysis of input-output relations. Here, model competence is defined as the internal, generalizable knowledge of a model that extends beyond its observable performance.
    Output-focused metrics, such as exact match accuracy or Levenshtein edit distance \cite{}, do not provide an accurate representation of a model's competence, only its performance on a specific task suite. Recent work demonstrate increasing interest in model competence analysis, analyzing models in domains such as spatial analysis \cite{zhang2025visionlanguagemodelsrepresentspace} and logical reasoning \citep{wang2025logicalformscomplementprobability}\hala{what other paper does Yixuan mention?}.
    As such, \toolname offers a toolkit for localizing knowledge and the causal hidden mechanisms in VLMs, which enables model competence analysis beyond simply accuracy-based performance evaluations.
\fi

\vspace{2pt}\noindent\textbf{Transformer interpretability toolkits.}
With increasing interest in interpreting Transformers, there have been various toolkits supporting their interpretation and analysis \interalia{clark2019does,nanda2022transformerlens,belrose2023elicitinglatentpredictionstransformers,ali2025entropylensinformationsignaturetransformer}.
However, most existing generic toolkits focus on text-only \citep[e.g., ][]{nanda2022transformerlens} or vision-only \citep[e.g.,][]{joseph2025prisma} Transformers, while the VLM counterparts significantly fall behind.\footnote{For example, TransformerLens \citep{nanda2022transformerlens} can be patched to support vision analysis, but requires a non-trivial setup and is rigid in extending functionality.}
To the best of our knowledge, all existing VLM interpretability toolkits \citep{palit2023towards,ben2024lvlm,neo2024towards} support no more than a single model, and the designs are not easily extensible to support other models.
To bridge this gap, \toolname provides a unified framework to extract internal representations of VLMs, which can be coupled with existing interpretability methods \interalia{thrush2022winoground,zhang2024large,basu2024understanding,stevens2025sparse} to analyze and assess a wide range of state-of-the-art VLMs.

\section{VLM-Lens}

At a high level, the \toolname toolkit incorporates the hook mechanism of PyTorch \citep{paszke2019pytorch} to extract the internal representations of VLMs, which are then stored in a database for further analysis.
We detail the design philosophy and key implementation aspects as follows.

\vspace{2pt}\noindent\textbf{Design philosophy.}
The design of \toolname is driven by the need for a simple yet flexible toolkit to extract internal representations while accommodating the diverse dependency requirements of different VLMs.
This is implemented through a central interface (\texttt{\href{https://github.com/compling-wat/vlm-lens/blob/main/src/main.py}{src/main.py}}) that accepts model-specific configuration files in YAML format, along with dedicated environment setups for each model.
Each model-specific implementation inherits from the base class, which standardizes model loading, preprocessing, and inference, while still allowing for model-specific customizations.
The extracted intermediate representations are stored in a database with a standardized schema, enabling efficient retrieval and analysis.
This design, executed by rigorous peer code review, ensures high extensibility: to support a new model, developers only need to implement a new model-specific class without modifying the core logic of the toolkit.

\vspace{2pt}\noindent\textbf{Supported VLMs.}
We currently support: Aya-Vision \citep{dash2025aya}, Blip-2 \cite{li2023blip}, CLIP \citep{radford2021learning}, CogVLM \cite{wang2024cogvlm}, GLaMM \citep{rasheed2024glamm}, InternLM-XComposer \cite{zhang2023internlm}, InternVL \cite{chen2024internvl}, Janus \citep{wu2025janus}, LLaVA \cite{liu2023visual}, MiniCPM-o \citep{team2025minicpm} and MiniCPM-V-2 \cite{yao2024minicpm}, Molmo \cite{deitke2024molmo}, Paligemma \cite{beyer2024paligemma}, Pixtral \citep{agrawal2024pixtral}, PerceptionLM \citep{cho2025perceptionlm} and Qwen2-VL \cite{wang2024qwen2}.
The toolkit supports all variants of these models across sizes, with the only requirement being sufficient hardware resources to load the model.

\vspace{2pt}\noindent\textbf{Configuration files.}
\toolname allows users to specify model configurations, input and output data paths, model layers of interest, and other model-specific parameters through YAML configuration files.
Users may extend the configuration files to include additional parameters for their experiments.
No hard-coded parameters are used throughout the codebase.
As an example, the following file (\texttt{\href{https://github.com/compling-wat/vlm-lens/blob/main/configs/models/blip2/blip2.yaml}{configs/models/blip2/blip2.yaml}}) specifies the required parameters for extracting the output of layers \texttt{vision\_model.post\_layernorm} and \texttt{language\_model.lm\_head} in \texttt{Blip2-OPT-2.7B} \citep{li2023blip}:
\begin{figure}[H]
    \centering
    \vspace{-5pt}
    \begin{minted}[fontsize=\scriptsize, frame=lines]{yaml}

architecture: blip2
model_path: Salesforce/blip2-opt-2.7b
model:
  - torch_dtype: auto
output_db: output/blip2.db
input_dir: ./data/test-images/
prompt: "Describe the color in this image in one word."
modules:
  - language_model.lm_head
  - vision_model.post_layernorm

        \end{minted}
    \vspace{-20pt}
\end{figure}
\noindent The command \verb|python src/main.py --config| \verb|configs/models/blip2/blip2.yaml| will provide the user with the corresponding intermediate output tensors in \texttt{output/blip2.db}, a SQLite3 database that can be further queried for analysis.

\vspace{2pt}\noindent\textbf{Pipeline implementation and output database organization.}
To initialize a model, an accompanying preprocessor is required to process the input images and format the prompt: for example, the chat template for LLaVA-1.5-7B \cite{liu2023visual} and Qwen2-VL \cite{wang2024qwen2} is implemented in the preprocessor.
The hidden representation extraction process involves registering a forward hook, a callable function that provides access to the input and output tensors of a layer specified by the user.
It takes the preprocessed image and prompt as input, and saves the output tensors in a database following the forward pass.
All forward hooks are unregistered to prevent contamination in later iterations.
A list of possible layers (i.e., modules in PyTorch) of a particular model can be returned easily using the \verb|--log-named-modules| parser argument, if the user requires more information on a model's internal structure.
Lastly, the extracted tensors are stored in a database with the following attributes: \texttt{name, architecture, image\_path, prompt, label, layer, tensor\_dim, tensor}. These attributes correspond to the model HuggingFace identifier (e.g., \texttt{Salesforce/blip2-opt-2.7b}), model architecture specified in this toolkit (e.g., \texttt{blip2}), image path, textual prompt content, label of the example (if applicable), layer name (e.g., \texttt{language\_model.lm\_head}), the dimensionality of the extracted tensor, and the extracted intermediate result tensor itself, respectively.

While we provide a default preprocessor and a hooked forward pass implementation in \texttt{\href{https://github.com/compling-wat/vlm-lens/blob/main/src/models/base.py}{src/models/base.py}} that can handle simple cases (e.g., LLaVA), these functions can be overridden to accommodate model-specific requirements (see more examples in \texttt{\href{https://github.com/compling-wat/vlm-lens/blob/main/src/models}{src/models}}).

\vspace{-5pt}
\section{Usage Example I: Probing}
\label{sec:demo}

\vspace{-5pt}
\subsection{Experimental Setups}
We first demonstrate the usage of \toolname by probing \citep{ettinger-etal-2016-probing}, where we train a set of probes on the extracted representations, evaluating the internal competence of VLMs on recognizing a set of primitive concepts.

\vspace{2pt}\noindent\textbf{Dataset.}
We create our dataset using CLEVR \citep{johnson2017clevr}.
As depicted in \cref{table:clevr-example}, we define five categorical splits: \texttt{color}, \texttt{material}, \texttt{number}, \texttt{shape}, and \texttt{size}, each corresponding to a primitive object attribute in the images, as well as a \texttt{boolean} split that may cover any attribute with binary questions.
Each split can be considered as a $c$-way classification task, where $c$ represents the number of possible choices in the split. 
For each split, 80\% data is used for training the probe, whereas the remaining 20\% is used for testing.\footnote{Data available at \url{https://huggingface.co/compling}.}

\begin{table}[t!]
    \centering
    \includegraphics[width=1.0\linewidth]{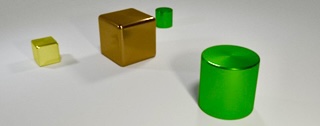} \\[3pt]
    \scalebox{0.9}{
    \begingroup
    \renewcommand{\arraystretch}{0.8}
    \setlength{\tabcolsep}{3.75pt}
    \hspace{-12pt}
    \begin{tabular}{lp{0.65\linewidth}r}
        \toprule
        \textbf{Split}    & \textbf{Question}                                                                                              & \textbf{Answer} \\
        \midrule
        \texttt{boolean}  & \setlength{\baselineskip}{12pt}Are there any other things that are the same size as the brown object?                                         & yes             \\[4pt]
                \cmidrule(lr){1-3}

        \texttt{color}    & \setlength{\baselineskip}{12pt}There is a small cylinder that is made of the same material as the big brown thing; what color is it?
                          & green                                                                                                                            \\[4pt]
        \cmidrule(lr){1-3}
        \texttt{material} & \setlength{\baselineskip}{12pt}What is the material of the big object that is the same color as the small metal cylinder?                     & metal           \\[4pt]
        \cmidrule(lr){1-3}
        \texttt{number}   & \setlength{\baselineskip}{12pt}What number of things are brown blocks or green metallic cylinders that are to the right of the tiny cylinder?
                          & 2
        \\[4pt]
        \cmidrule(lr){1-3}
        \cmidrule(lr){1-3}
        \texttt{shape}    & \setlength{\baselineskip}{12pt}The big brown shiny thing is in what shape?                                                                    & cube
        \\[4pt]
        \cmidrule(lr){1-3}
        \texttt{size}     & \setlength{\baselineskip}{12pt}There is another thing that is the same shape as the brown metallic object; what is its size?                  & small           \\[4pt]
        \bottomrule
    \end{tabular}
    \endgroup}
    \vspace{-6pt}
    \caption{
        \label{table:clevr-example}
        Examples from our probing dataset. 
        We input the image and the question to the VLM, extract hidden states, and train a lightweight probe to decode the answer from these representations.
        \vspace{-15pt}
    }
\end{table}
\begin{figure}[t!]
    \centering
    \includegraphics[width=0.99\linewidth]{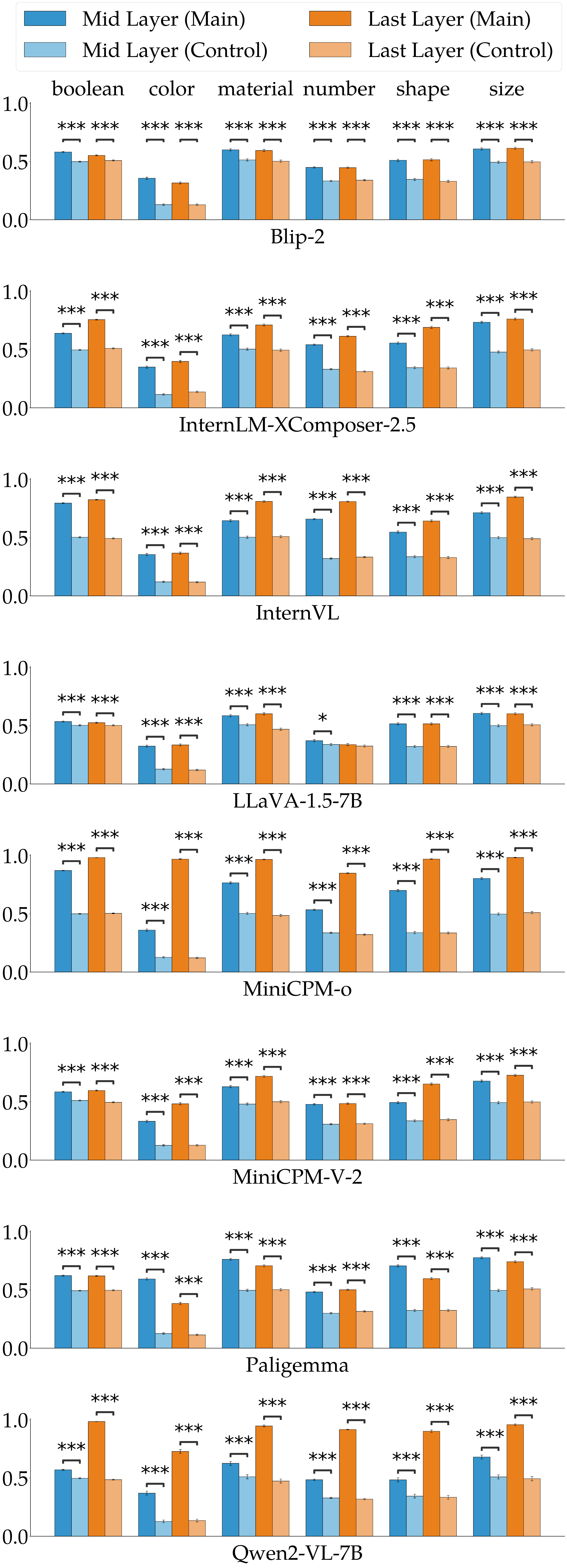}
    \caption{
    Evaluation Accuracy on our probing dataset by model, layer, and split. 
    \textit{Main} refers to probing on the regular data, while \textit{control} stands for probing using data with random labels. 
    The number of asterisks represents the significance level of the Z-test for Bernoulli variables (***: $p=.001$, **: $p=.01$, *: $p=.05$). 
    }
    \label{fig:clevr-results}
\end{figure}

\vspace{2pt}\noindent\textbf{Probing approach.} 
Each combination of split $D$, model $\mathcal{M}$, and layer $\ell$ is considered independently in the probing process. 
For an example $(e_i, y_i) \in D$, where $e_i$ represents the input (i.e., image and text) of the task and $y_i$ stands for its ground-truth prediction category, we extract the intermediate output of model $\mathcal{M}$ at the layer $\ell$ with mean pooling across the input tokens, denoted as $\mathcal{M}_\ell (e_i) \in \mathbb{R}^d$, where $d$ is the dimensionality of layer $\ell$.
$\mathcal{M}_\ell (e_i)$ is used as the feature of $e_i$ to train a probe that best predicts the $y_i$ label.  

We employ a two-layer perceptron as the probe, with ReLU activations and 512 hidden units.\footnote{We searched for the probe hidden size with small-scale experiments, and found 512 to be the hidden size with the best performance consistency.}
A $k$-fold cross-validation is conducted to search for the best optimization hyperparameters for each split, using matching-based accuracy as the metric. 
We then train the probe model on the full training set using the best optimization hyperparameters, and report the accuracy on the test set. 
Following \citet{hewitt2019designinginterpretingprobescontrol}, we complement our main probe (trained to predict the true ground-truth labels) with a control probe (trained on randomly shuffled labels). 
A statistically significant advantage of the main probe over the control probe suggests that the VLM encodes task-relevant information, whereas no advantage implies that the probe relies on its own capacity to memorize spurious patterns.

\vspace{-5pt}
\subsection{Results and Discussion}

We evaluate the middle (i.e., $\lceil\frac L2\rceil$) and last layers of eight supported models, where $L$ is the total number of layers in the model of interest (\cref{fig:clevr-results}).
Notably, the probes trained on the Qwen-7b and MiniCPM-o representations achieve an almost perfect accuracy within many dataset splits, with a statistically significant difference from the control probe performance. 
This effect is especially prevalent in the results of the last layer representations with a few exceptions (e.g., Blip-2 and Paligemma). 
In contrast, models like LLaVA-1.5-7B demonstrate a minor (albeit statistically significant in most cases) difference between the evaluation results in the original and control conditions, indicating weaker competence.

Across all models, the \texttt{color} attribute appears to be the most salient feature, with the main probe performance significantly better than the control results across both layers. 
Models with instruction-following and multimodal understanding capabilities, such as Qwen2-VL, MiniCPM-o, and InternVL, performed well on the more difficult splits such as \texttt{material}, \texttt{number}, and \texttt{shape}, especially when using the last-layer representations. 
This probing-based competence evaluation complements existing accuracy-driven benchmarks by providing a more detailed understanding of what is represented in the internal states, grounding the model performance in interpretable primitive knowledge.

\section{Usage Example II: Concept Similarity}

\begin{figure*}[!t]
    \centering
    \begin{subfigure}[t]{.32\textwidth}
        \centering
        \includegraphics[width=1.05\columnwidth]{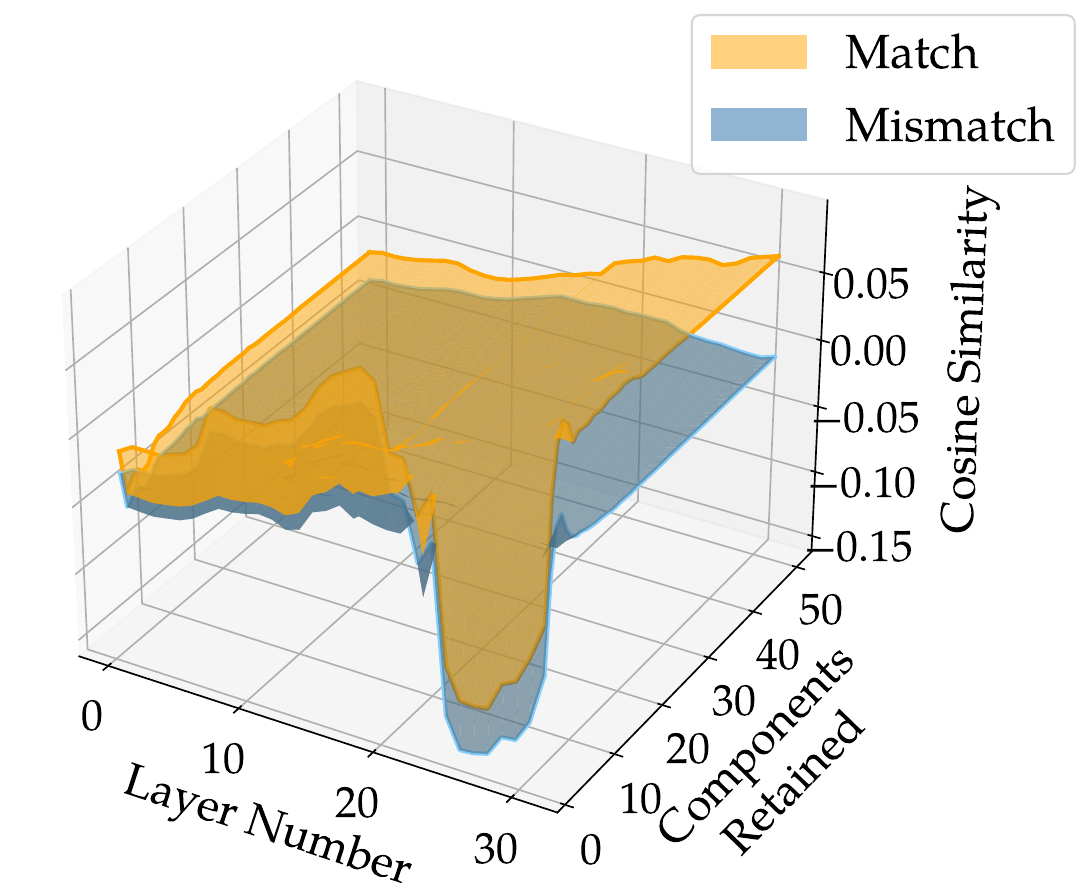}
        \vspace*{-15pt}
        \caption{Lexical content vs. reference.}
        \label{fig:llava-color-txt}
    \end{subfigure}
    ~
    \begin{subfigure}[t]{.32\textwidth}
        \centering
        \includegraphics[width=1.05\columnwidth]{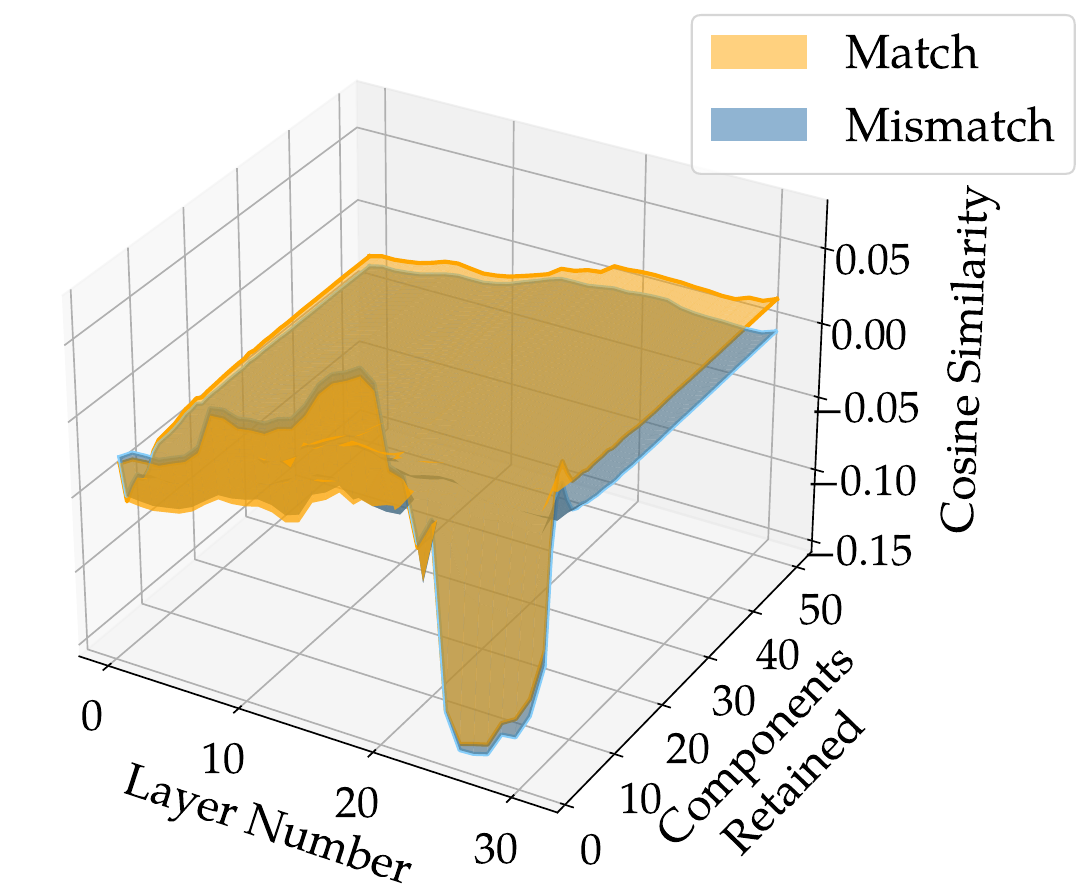}
        \vspace*{-15pt}
        \caption{Foreground font color vs. reference.}
        \label{fig:llava-color-fg}
    \end{subfigure}
    ~
    \begin{subfigure}[t]{.32\textwidth}
        \centering
        \includegraphics[width=1.05\columnwidth]{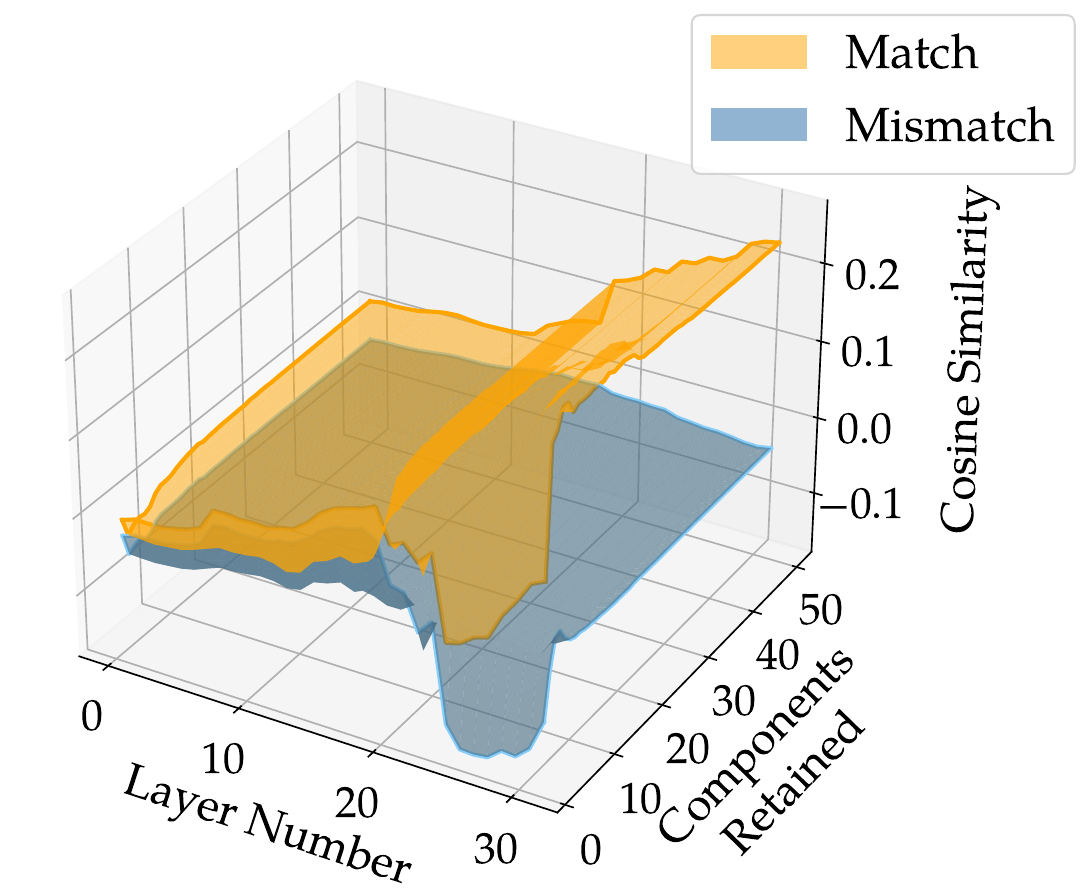}
        \vspace*{-15pt}
        \caption{Background color vs. reference.}
        \label{fig:llava-color-bg}
    \end{subfigure}
    \vspace*{-5pt}
    \caption{Cosine similarity between Stroop task images and primitive color concepts. 
    Results are shown as a function of model layer (x-axis) and number of PCA components retained (y-axis), with orange surfaces indicating matching conditions and blue surfaces indicating mismatching conditions when considering different aspects. \vspace*{-10pt}}
    \label{fig:llava-color}
\end{figure*}

\begin{figure}
    \centering
    \begin{subfigure}[t]{.45\linewidth}
        \includegraphics[width=\linewidth]{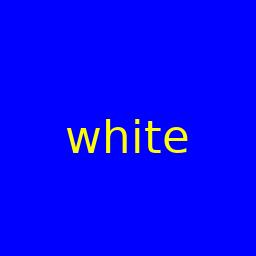}
        \caption{Word \textit{white} written in yellow on blue canvas.}
    \end{subfigure}
    \hspace{10pt}
    \begin{subfigure}[t]{.45\linewidth}
        \includegraphics[width=\linewidth]{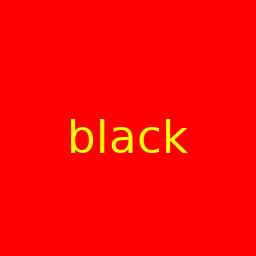}
        \caption{Word \textit{black} written in yellow on red canvas.}
    \end{subfigure}
    \vspace{-8pt}
    \caption{Example images used in the Stroop Task. 
    \vspace{-10pt}}
    \label{fig:stroop}
\end{figure}

The second experiment is inspired by the classic \textit{Stroop Effect}~\citep{stroop1935studies}, which demonstrates that humans exhibit slower and more error-prone responses when asked to name the font color of a word that is itself an incongruent color term (e.g., the word red printed in blue ink). 
We adapt this paradigm to VLMs by constructing images with deliberate incongruities between three color cues (\cref{fig:stroop}): the lexical word (e.g., white), the font color (e.g., yellow), and the background color (e.g., blue). 
This design probes how VLMs ground the notion of color under ambiguous instructions.

\subsection{Experimental Setups}
\vspace{2pt}\noindent\textbf{Prompt setup.}
Unlike humans in the Stroop task, who are explicitly instructed to name the ink color, we query the model with an intentionally ambiguous prompt: \textit{Describe the color in this image in one word.}, coupled with a single image with incongruent cues.
This allows us to study which representation (lexical, font, or background) the model privileges in its internal embeddings.

\vspace{2pt}\noindent\textbf{Prototype construction.}
To establish references for primitive color concepts (e.g., red, blue, green), we retrieve the  top 10 de-duplicated Creative Commons--licensed images using the Google Images API.\footnote{\url{https://developers.google.com/photos}}
We extract the intermediate layer representations of these images coupled with the ambiguous textual instruction using \toolname.
% Users can generate their own concept prototypes and evaluate them relative to target images.

\vspace{2pt}\noindent\textbf{Similarity-based analysis.}
We apply Principal Component Analysis (PCA) to identify the directions that capture the most color variation. 
Let the zero-meaned embeddings be denoted as 
$E \in \mathbb{R}^{n \times d}$,
where $n$ is the number of reference images corresponding to primitive color concepts, and $d$ is the dimensionality of the intermediate layer output. PCA learns a linear projection $W \in \mathbb{R}^{d\times d'}$ that maps $E$ to a lower-dimensional representation $E' = EW \in \mathbb{R}^{n \times d'} (d' \ll d)$.

We evaluate the Stroop task images under different numbers of retained principal components: for one image at a specific model layer, we extract the corresponding hidden representation, projecting it into the transformed space with $W$, and compute the average cosine similarity with the reference color concepts. 
For lexical, font, and background colors, we report the average matched and mismatched cosine similarities across layers.

\subsection{Results and Discussion}
We visualize the results for LLaVA-1.5-7B (\cref{fig:llava-color}), where a larger gap between matched and mismatched data indicates a more prominent feature in the Stroop task.
We find a clear separation between the match and mismatch conditions across all three settings, indicating all three types of information (i.e., lexical content, foreground font color, and background color) are reliably encoded in the model.
As expected, background color (\cref{fig:llava-color-bg}) produces the strongest contrast between matched and mismatched examples.
However, somewhat surprisingly, color presented as lexical content is more prominent than font color, evidenced by the gap in \cref{fig:llava-color-txt} than that in \cref{fig:llava-color-fg}.
In addition, all concepts require a sufficient number of PCA components to achieve a clear separation, suggesting that color information is not captured in a single linear direction in the representation space.

\section{System Evaluation}

\begin{table*}[t]
    \centering
    \scalebox{0.9}{
    \begingroup
    \renewcommand{\arraystretch}{0.9}
    \setlength{\tabcolsep}{5.2pt}
    \hspace{-10pt}
    \begin{tabular}{lrrrrr}
        \toprule
        \multirow{2}{*}{\textbf{Model}} & \multirow{2}{*}{\textbf{\# Params}} & \multirow{2}{*}{\textbf{Precision}} & \textbf{Peak Mem} & \textbf{Inference Time} & \textbf{Per-Instance Time} \\
        &                                     &                                      & (MB)              & (seconds)              & (seconds)                        \\
        \midrule
        CLIP~\shortcite{radford2021learning}                   & 150M               & \texttt{float32}   & 616.69   & 135 & 0.025 \\
        Blip-2~\shortcite{li2023blip}                 & 2.7B               & \texttt{float32}   & 15,261.28 & 295 & 0.055 \\
        % CogVLM & 10B & && 815 & 0.076 \\
        % GlaMM & 7B & \texttt{float32} & 14,147.85 & 1091 & 0.203 \\
        InternLM-XComposer-2.5~\shortcite{zhang2023internlm} & 7B                 & \texttt{bfloat16}  & 24,037.64 & 3,056 & 0.569 \\
        InternVL~\shortcite{chen2024internvl}               & 8B                 & \texttt{bfloat16}  & 21,136.92 & 3,347 & 0.623 \\
        LLaVA-1.5~\shortcite{liu2023visual}              & 7B                 & \texttt{float16}   & 29,031.55 & 1,566 & 0.291 \\
        MiniCPM-V-2~\shortcite{yao2024minicpm}            & 2.8B               & \texttt{bfloat16}  & 7,154.23 & 495 & 0.092 \\
        MiniCPM-o~\shortcite{team2025minicpm}              & 8B                 & \texttt{bfloat16}  & 18,058.20 & 671 & 0.125 \\
        Molmo~\shortcite{deitke2024molmo}                  & 7B                 & \texttt{float32}   & 34,403.26 & 2,841 & 0.529 \\
        Paligemma~\shortcite{beyer2024paligemma}              & 3B                 & \texttt{float32}   & 12,018.30 & 554 & 0.103 \\
        Qwen2-VL~\shortcite{wang2024qwen2}               & 7B                 & \texttt{bfloat16}  & 33,840.66 & 1,497 & 0.279 \\
        \bottomrule
    \end{tabular}
    \endgroup}
    \caption{Inference properties of different models in \toolname after execution on the MSCOCO \cite{lin2014microsoft} dataset, including GPU memory usage and inference time metrics. \vspace{-10pt}}
    \label{tab:inf-data}
\end{table*}

We evaluate \toolname, in terms of time and memory efficiency, on the inference procedure of our supported models, using a subset of the MSCOCO validation set \citep{lin2014microsoft}\footnote{\url{https://huggingface.co/datasets/compling/coco-val2017-obj-qa-categories}} with 2,690 examples (\cref{tab:inf-data}).
All experiments are done on a single NVIDIA-A40 GPU with sufficient CPU memory, using an inference batch size of 1.
Users may use the reported performance statistics for informed choices on GPU selection and advanced inference techniques. 

\subsection{Inference Time}

The model inference times are calculated for the duration of the forward inference on the dataset, including the database saving execution, and disregarding the model and processor loading time. 
As expected, CLIP is the fastest despite its high precision point, which is a result of its small number of parameters.
It is worth noting that CLIP is also the only model trained using contrastive image-text matching, rather than incorporating visual tokens alongside text, and, therefore, is not directly comparable to other models.
InternVL, InternLM-XComposer-2.5, and MolMo are the slowest models, with high per-inference times compared to other models with similar parameter counts (7B). 
Furthermore, within the same parameter count and precision point, InternVL and MiniCPM-o demonstrate a disparate difference in inference time (0.623s vs 0.125s per-inference, respectively), which is likely due to differing architecture optimizations and input processing methods.

\subsection{Memory Usage}

To approximate the memory usage of each model using our toolkit, we record the precision and peak GPU memory usage (in MB; Table \ref{tab:inf-data}). 
Similar to inference time, CLIP demonstrates the lowest memory footprint (617MB), due to its compact architecture. 
In contrast, LLaVA-1.5 and Qwen2-VL use the most memory, despite having a low precision point and the same parameter count as many other models. 
In general, \texttt{bfloat16} and \texttt{float16} precision points reduce memory usage, but their effectiveness varies depending on the architecture: InternLM-XComposer and Qwen2-VL still demonstrate a large memory footprint ($>$24GB).

\section{Conclusion and Discussion}
In this paper, we introduce \toolname, a toolkit that aims to benchmark, analyze, and interpret VLMs systematically.
We demonstrate that the toolkit enables the assessment of the internal competence of a wide range of VLMs (\cref{sec:demo}), going beyond the simple accuracy-based evaluations provided by most existing benchmarks \interalia{yue2024mmmumassivemultidisciplinemultimodal}.
We offer performance statistics (\cref{tab:inf-data}) for users' informed choice on both models and hardware. 
Users who develop new VLMs through training and fine-tuning can also easily use the provided probing framework to diagnose their model capabilities.
We anticipate that this toolkit will lower the barrier for evaluations of VLMs in a scientifically rigorous way.

Results from our demonstrative experiment (\cref{sec:demo}) align with the caveat that performance alone may be insufficient for evaluating models \citep{hu2023prompting,zhang2025visionlanguagemodelsrepresentspace,wang2025logicalformscomplementprobability}. 
Although all evaluated models are considered highly capable on existing benchmarks, many still fail on simple synthetic data. 
These findings reaffirm concerns about the reliability of VLMs, and \toolname will actively support their analysis and improvement along these lines. 

\vspace{2pt}\noindent\textbf{Limitations and planned community support.}
The current toolkit does not directly support more downstream tasks than probing, such as attention interpretation and neural circuit discovery. 
Additionally, our current inference and database storage approach prevents the use of gradient-based saliency analyses such as Grad-CAM \citep{selvaraju2017grad}. 
Current users still need to implement their customized functions; however, we anticipate rapid community contributions to the repository for a diverse range of tasks---efforts we are committed to supporting in the long term.
\section*{Acknowledgments}
We thank Yifan Jiang and Michael Ogezi for their constructive feedback.
This work is supported in part by NSERC RGPIN-2024-04395, a Vector Scholarship to HS, and a Canada CIFAR AI Chair award to FS.

\bibliography{custom}
\bibliographystyle{acl_natbib}

\newpage
\appendix
\section{Contribution Statement}
We list the contributions of each author below:
\begin{itemize}[leftmargin=*, noitemsep, topsep=0pt]
    \item Project idea: Freda Shi, Ziqiao Ma.
    \item Substantial prototype design: Eric Huang, Hala Sheta, Freda Shi.
    \item Prototype implementation: Eric Huang.
    \item Model implementation: Shuyu Wu, Jiajun Hong, Hala Sheta, Eric Huang, Ilia Alenabi, Ryker Lin, Ruoxi Ning, Daniel Wei, Jialin Yang.
    \item Documentation, user guide, and tutorials: Ziqiao Ma, Freda Shi, Ilia Alenabi, Ryker Lin, Daniel Wei.
    \item Substantial code review: Hala Sheta, Shuyu Wu, Eric Huang, Freda Shi.
    \item Probing experiments: Hala Sheta, Shuyu Wu, Ilia Alenabi, Jiajun Hong, Ryker Lin, Daniel Wei.
    \item Concept similarity experiments: Ziqiao Ma, Shuyu Wu, Freda Shi.
    \item Substantial project discussion: Hala Sheta, Eric Huang, Shuyu Wu, Jiajun Hong, Jiawei Zhou, Ziqiao Ma, Freda Shi.
    \item Paper writing: Hala Sheta, Freda Shi, Ziqiao Ma.
    \item Project supervision: Freda Shi, Jiawei Zhou, Ziqiao Ma.
\end{itemize}

\section{Probing Configuration} \label{appendix:probe}
For each of our supported models, we extract the hidden representations of the middle (e.g., 16) and last (e.g., 32) of the \texttt{post-attention layer norm} module (or equivalent) using our library. This is because we require a 4096-dimensional (or equivalent) representation that can be efficiently stored and contains some relevant information about the input.

Each probe is instantiated with \texttt{ReLU} activation and a hidden size of 512, selected empirically to balance performance and efficiency. The input and output sizes are determined dynamically based on the tensors and labels in the input database. We tune the hyperparameters, \texttt{learning\_rate}, \texttt{num\_epochs} and \texttt{batch\_size}, to find the best combinations that incurs the lowest mean validation loss after $k$-fold cross-validation with $k=5$. With 3 options for each former parameter, we search through $3^3 = 27$ configurations. Lastly, using the best training configuration, we train two new models from scratch on the original and shuffled datasets, respectively.

\end{document}